\documentclass[runningheads]{llncs}
\usepackage{graphicx}
\usepackage[hidelinks]{hyperref}

\usepackage{multirow}
\usepackage[dvipsnames]{xcolor}

\usepackage[dvipsnames]{xcolor}
\usepackage{tikz}

\usetikzlibrary{shapes,shadows,arrows,backgrounds}
\tikzstyle{line} = [draw, -stealth, thick]
\tikzstyle{arrow} = [thick,->,>=stealth]
\tikzstyle{inputIMG} = [draw, rectangle, fill=Gray!25, text width=15mm, text centered, minimum height=15mm, minimum width=15mm]
\tikzstyle{input} = [draw, rectangle, fill=Gray!25, text width=15mm, text centered, minimum height=7mm, minimum width=15mm]
\tikzstyle{block} = [draw, rectangle, text width=15mm, text centered, minimum height=10mm, node distance=6mm]
\tikzstyle{blockCNN} = [draw, trapezium, fill=SeaGreen!25, text width=15mm, text centered, minimum height=10mm]
\tikzstyle{blockSentEnc} = [draw, rectangle, fill=YellowGreen!25, text width=15mm, text centered, minimum height=7mm, minimum width=15mm]
\tikzstyle{blockAtt} = [draw, rectangle, fill=orange!25, text width=20mm, text centered, minimum height=10mm, node distance=6mm]
\tikzstyle{block_small} = [draw, rectangle, text width=13mm, text centered, minimum height=3mm, node distance=6mm]
\tikzstyle{blockOCR_small} = [draw, rectangle, fill=blue!25, text width=13mm, text centered, minimum height=3mm, node distance=6mm]
\tikzstyle{blockAggr_small} = [draw, rectangle, fill=red!25, text width=13mm, text centered, minimum height=3mm, node distance=6mm]
\tikzstyle{blockAtt_small} = [draw, rectangle, fill=orange!25, text width=15mm, text centered, minimum height=3mm, node distance=6mm]
\tikzstyle{label} = [node distance=5mm]

\newcommand{\etal}{\textit{et al}. }

\usepackage{ bbold }

\begin{document}
%
%\title{A Comparative Analysis of Near-Duplicate Image Detection Techniques for Scanned Photo Collections}
\title{Transductive Learning for Near-Duplicate Image Detection in Scanned Photo Collections}
%
%\titlerunning{An analysis of Near-Duplicate Image Techniques for Photo Collections}
\titlerunning{Transductive Learning for Near-Duplicate Image Detection}
%\author{Francesc Net\orcidID{0000-0001-6888-519X} \and Lluis Gómez\orcidID{0000-0003-1408-9803}}

\author{Francesc Net\inst{1} \and Marc Folia\inst{2} \and Pep Casals\inst{2} \and Lluis Gómez\inst{1}}

\authorrunning{F. Net et al.}
\institute{
Computer Vision Center, Universitat Autònoma de Barcelona, Catalunya\\
\email{\{fnet,lgomez\}@cvc.uab.cat} \and Nubilum, Gran Via de les Corts Catalanes 575, 1r 1a | 08011 Barcelona \\ \email{\{pep.casals,marc.folia\}@nubilum.es}}

\maketitle              % typeset the header of the contribution
\begin{abstract}
This paper presents a comparative study of near-duplicate image detection techniques in a real-world use case scenario, where a document management company is commissioned to manually annotate a collection of scanned photographs. Detecting duplicate and near-duplicate photographs can reduce the time spent on manual annotation by archivists. This real use case differs from laboratory settings as the deployment dataset is available in advance, allowing the use of transductive learning. We propose a transductive learning approach that leverages state-of-the-art deep learning architectures such as convolutional neural networks (CNNs) and Vision Transformers (ViTs). Our approach involves pre-training a deep neural network on a large dataset and then fine-tuning the network on the unlabeled target collection with self-supervised learning. The results show that the proposed approach outperforms the baseline methods in the task of near-duplicate image detection in the UKBench and an in-house private dataset.

\end{abstract}
\section{Introduction}

Historical archives are the storehouses of humankind's cultural memory, and the documents stored in them form a cultural testimony asset of incalculable value. From all the documents stored and managed in Historical Archives, photographs represent an ever-growing volume. This increase in volume has led to a critical need for efficient and effective methods for managing the data contained in these archives, particularly in terms of annotating them with useful metadata for indexing.

When a new document collection of scanned photographs arrives in an archive, archivists will typically need to perform certain manual tasks of classification, annotation, and indexing. Detecting duplicate and near-duplicate photographs can alleviate the burden on archivists by reducing the amount of time spent annotating the same image multiple times. This is where computer vision techniques can play a critical role in automating the process and improving the efficiency of archival work.

\begin{figure}
\begin{minipage}{\textwidth}
    \setcounter{footnote}{-1}
    \setcounter{mpfootnote}{1}
    \centering
    \includegraphics[width=\textwidth]{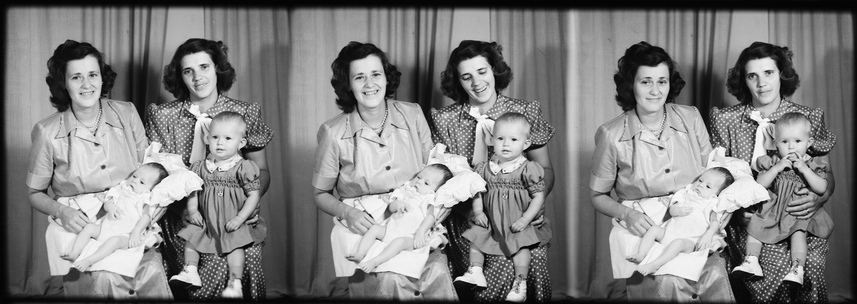}
    (a) \\
    \includegraphics[width=0.495\textwidth]{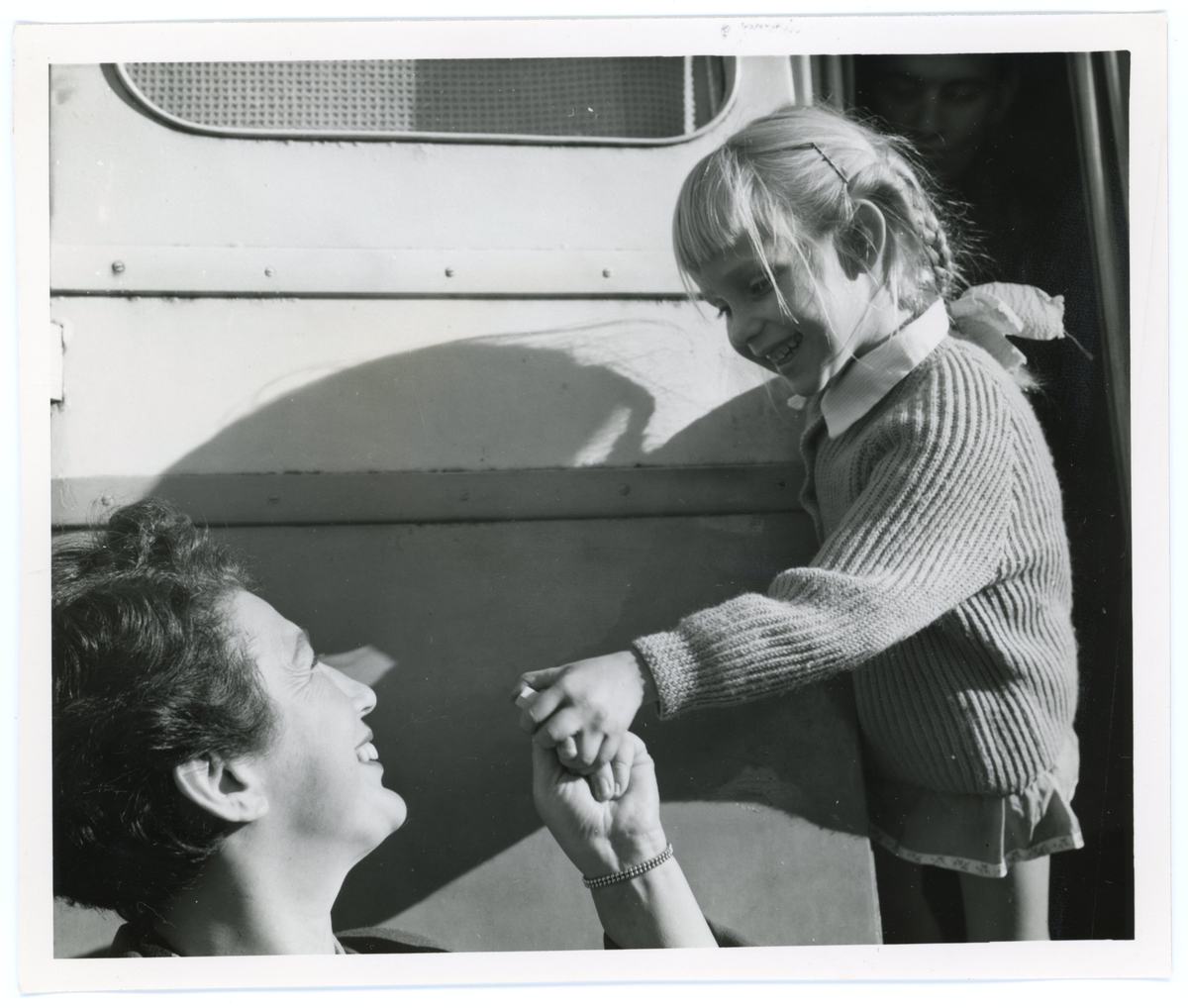} \includegraphics[width=0.495\textwidth]{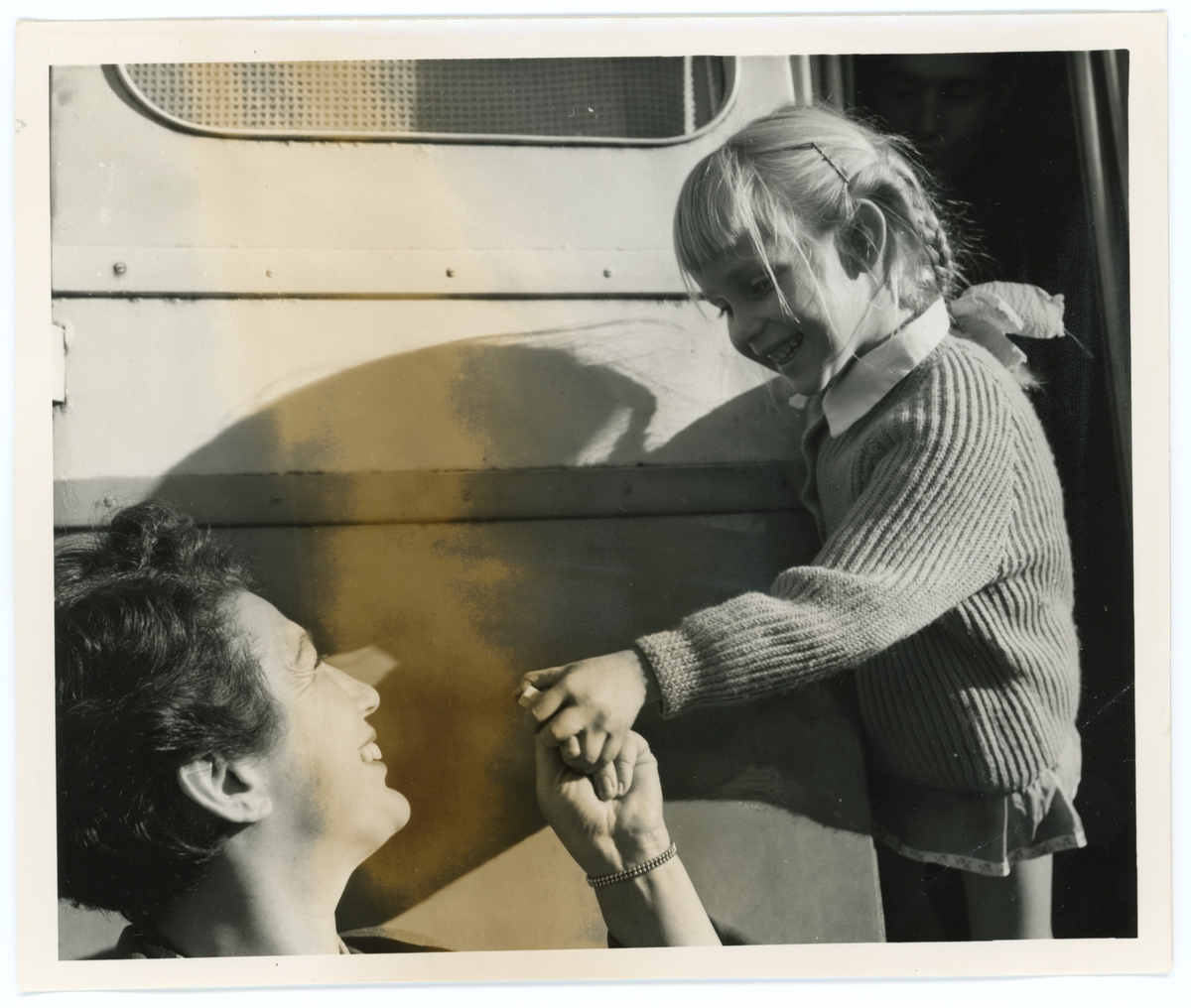}
    (b) \\
    \includegraphics[width=0.495\textwidth]{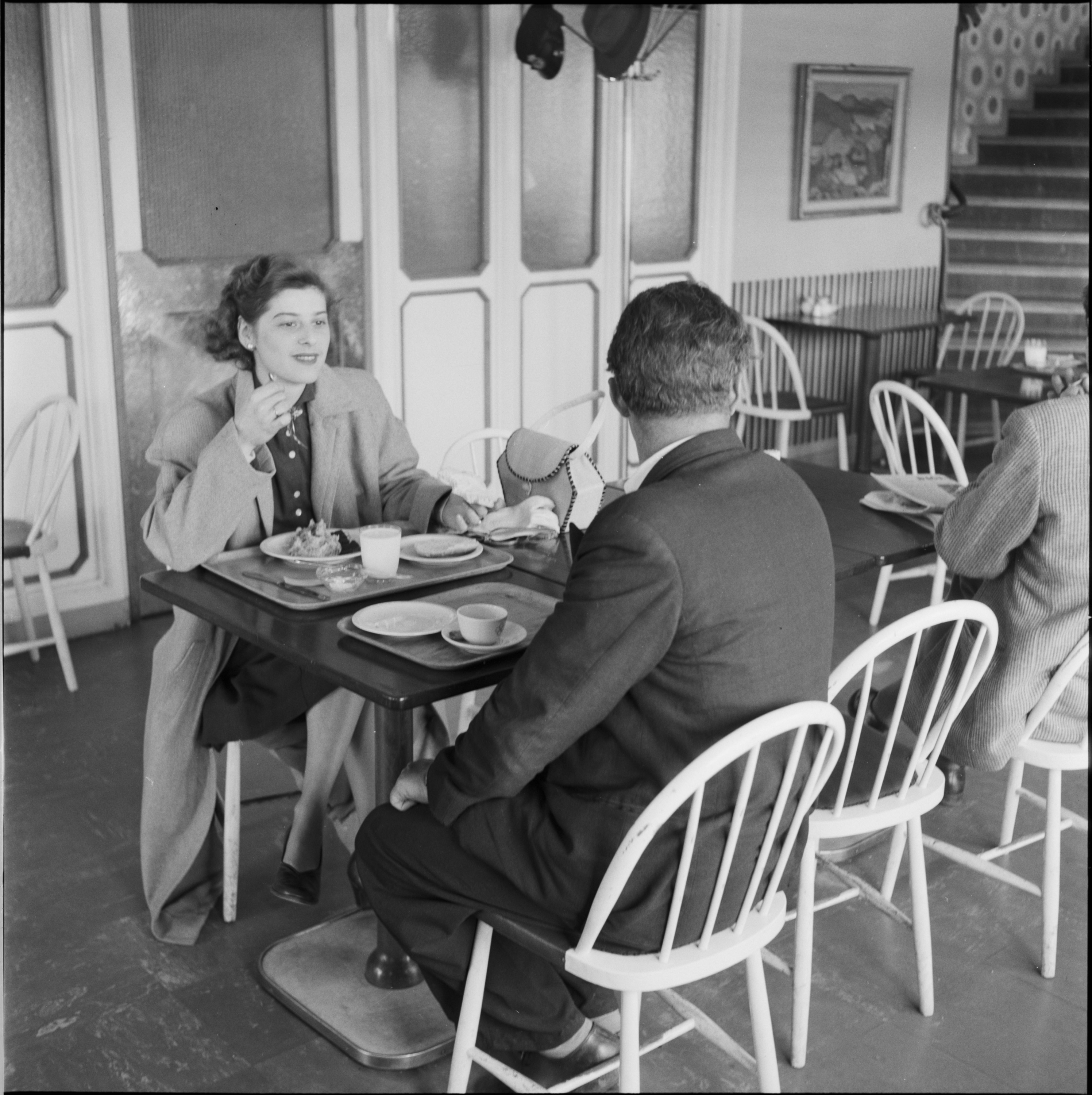} \includegraphics[width=0.495\textwidth]{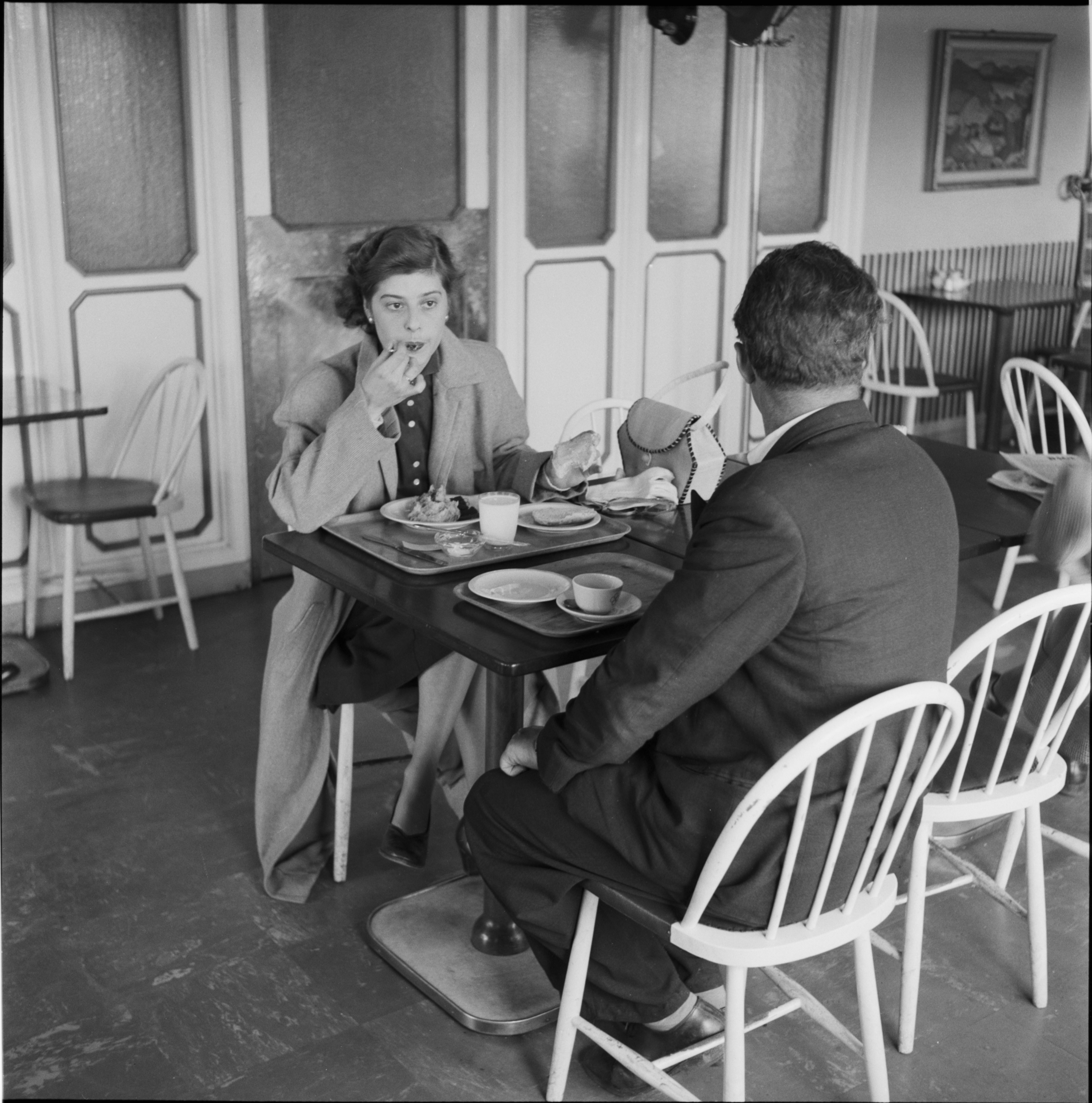}
    (c)
    \caption[Duplicate and near-duplicate images can appear in a given photo collection for different reasons: (a) shots from the same studio session, (b) scanned images from two paper originals with slight variations due to the paper wear over time, (c) different shots in a sequence of the same scene, etc. The images shown in this figure are reproduced from the DigitaltMuseum with Creative Common licenses (CC BY-NC-ND and CC pdm). The respective authors are: (a) Carl Johansson, (b) Anna Riwkin-Brick, and (c) Sune Sundahl.]{Duplicate and near-duplicate images can appear in a given photo collection for different reasons: (a) shots from the same studio session, (b) scanned images from two paper originals with slight variations due to the paper wear over time, (c) different shots in a sequence of the same scene, etc. The images shown in this figure are reproduced from the DigitaltMuseum\footnotemark ~with Creative Common licenses (CC BY-NC-ND and CC Public Domain). The respective authors are: (a) Carl Johansson, (b) Anna Riwkin-Brick, and (c) Sune Sundahl.}
    \footnotetext{\url{https://digitaltmuseum.se/}}
    \label{fig:teaser}
    \end{minipage}
\end{figure}

As illustrated in Figure~\ref{fig:teaser}, duplicate and near-duplicate images can appear in a given photo collection for different reasons and the number of them, while obviously varying for each collection, can represent a considerable percentage.

This paper presents a comparative study of various near-duplicate image detection techniques in the context of a real use case scenario. In particular, the study has been carried out in a document management company that has been commissioned to manually annotate a collection of scanned photographs from scratch. Furthermore, this real-world use case differs slightly from laboratory settings that are typically used to evaluate near-duplicate image detection techniques in the literature. Specifically, here the data set in which near-duplicate detection techniques are deployed is available in advance, which allows the use of transductive learning.

Transductive learning \cite{ref_transductive} is a type of learning that takes into account the information from the test data to improve the performance of the model. In contrast, in traditional inductive learning, the model is trained on a separate training set and then applied to a test set to evaluate its generalization capabilities.

In this paper, we consider fine-tuning a pre-trained model on the target collection with self-supervision as a form of transductive learning. The idea is that even though we do not have annotations on the test data set, we can use self-supervision techniques to learn representations that are useful for solving the near-duplicate detection task. In this way, as the model is adapted to the specific characteristics of the test data during the training process, this can result in improved performance compared to using a model previously trained on a different dataset.

Deep learning \cite{ref_dl_book} is a helpful and effective technique for a range of applications, but it also has drawbacks and limits when it comes to detecting near-duplicate images. One issue is that deep learning models must be trained using a significant quantity of labeled data. The model has to be trained on a broad and representative set of photos, including near-duplicate and non-duplicate images, in order to recognize near-duplicate images successfully. This procedure can take a lot of time and resources, especially if the data set is not yet accessible or if the photos are challenging to label appropriately.

Another issue is the possibility of overfitting, which occurs when a model performs well with training data but badly with fresh, untested data. In the case of recognizing near-duplicate photos, this might be especially troublesome, as the model may become extremely specialized in identifying the particular images on which it has been trained and may not generalize well to new images. 

Fortunately, in real applications like the one explained above, being able to access the collection of images gives us an advantage that we can exploit to mitigate these problems. Our focus in this article is to evaluate the different state-of-the-art vision architectures for near-duplicate image detection in these real use case scenarios. Our experiments show that the use of self-supervision techniques on the retrieval data set effectively improves their results. Although in our experiments we have used photographic collections, our study is also relevant for the detection of near duplicates in document images. In fact, some of the samples in the datasets we use contain textual information (scene/handwritten). 

The rest of the paper is organized as follows: the most recent state-of-the-art in near-duplicate detection is found in Section 2. In section 3, a list of the models employed is provided. In section 4, the experiments and the results are shown. Finally, section 5 contains the conclusions, discussions, and future work. 

\section{Related Work}
Applications like image search, copyright defense, and duplicate picture detection can all benefit from locating and classifying photos that are similar to or nearly identical to one another. The idea is to spot near-duplicate photographs and group them together so that just one image is maintained and the rest can be deleted or flagged for closer examination. Typically, the work is completed by comparing image properties, such as histograms, textures, or CNN features, and utilizing similarity metrics to gauge how similar the images are to one another. 

The evolution of near-duplicate image detection systems over the last two decades has been marked by a shift towards the use of deep learning techniques and an increasing emphasis on the development of more efficient and accurate methods for image feature extraction and matching. Before the use of deep learning, the primary approach was to use CBIR (Content-Based Image Retrieval) techniques. Chum \etal\cite{ref_lncs1_nd8}, for instance, makes use of an upgraded min-Hash methodology for retrieval and a visual vocabulary of vector quantized local feature descriptors (SIFT) to perform near-duplicate image detection. Similarly, Dong \etal\cite{ref_lncs1_nd9} use SIFT features with entropy-based filtering.

Thyagharajan \etal\cite{ref_lncs1_survey_nd} provides a survey of various techniques and methods for detecting near-duplicate images, including content-based, representation-based, and hashing-based methods and all of their variations. They discuss the issues and challenges that arise when dealing with near-duplicate detection, including the trade-off between precision and recall, approach scalability, and the impact of data distribution. 

Liong \etal\cite{ref_lncs1_nd4} and Haomiao \etal\cite{ref_lncs1_nd5} use a deep neural network to learn compact binary codes for images, by seeking multiple hierarchical non-linear transformations. The methods are evaluated in medium-scale datasets (e.g. CIFAR-10). Similarly, Dayan \etal\cite{ref_lncs1_nd6} and Fang \etal\cite{ref_lncs1_nd7} map images into binary codes while preserving multi-level semantic similarity in medium-scale multi-label image retrieval. Both methods are evaluated in the MIRFLICKR-25K dataset. 

Global and local CNN features can be used as in \cite{ref_lncs1_nd1} to detect near-duplicates. In order to filter out the majority of irrelevant photos, the method uses a coarse-to-fine feature matching scheme, where global CNN features are first created via sum-pooling on convolutional feature maps and then matched between a given query image and database images. On the other hand, Lia and Lambreti \etal\cite{ref_lncs1_nd2} discuss the performance of deep learning-based descriptors for unsupervised near-duplicate image detection. The method tests different descriptors on a range of datasets and benchmarks and compares their specificity using Receiver Operating Curve (ROC) analysis. The results suggest that fine-tuning deep convolutional networks are a better choice than using off-the-shelf features for this task. 

The applications of near-duplicate image detection can be very wide. He \etal\cite{ref_lncs1_nd10}, for instance, performs vehicle re-identification to distinguish different cars with nearly identical appearances by creating a custom CNN.

More recent publications \cite{ref_lncs1_nd3} use spatial transformers in comparison with convolutional neural networks (CNN) models. The method, which is designed to make better use of the correlation information between image pairs, utilizes a comparing CNN framework that is equipped with a cross-stream to fully learn the correlation information between image pairs. Additionally, to address the issue of local deformations caused by cropping, translation, scaling, and non-rigid transformations, the authors incorporate a spatial transformer module into the comparing CNN architecture. The method is tested on three popular benchmark datasets, California, Mir-Flickr Near Duplicate, and TNO Multi-band Image Data Collection, and the results show that the proposed method achieves superior performance compared to other state-of-the-art methods on both single-modality and cross-modality tasks.

\section{Methods}
\begin{figure}
    \centering
    \includegraphics[width=\textwidth]{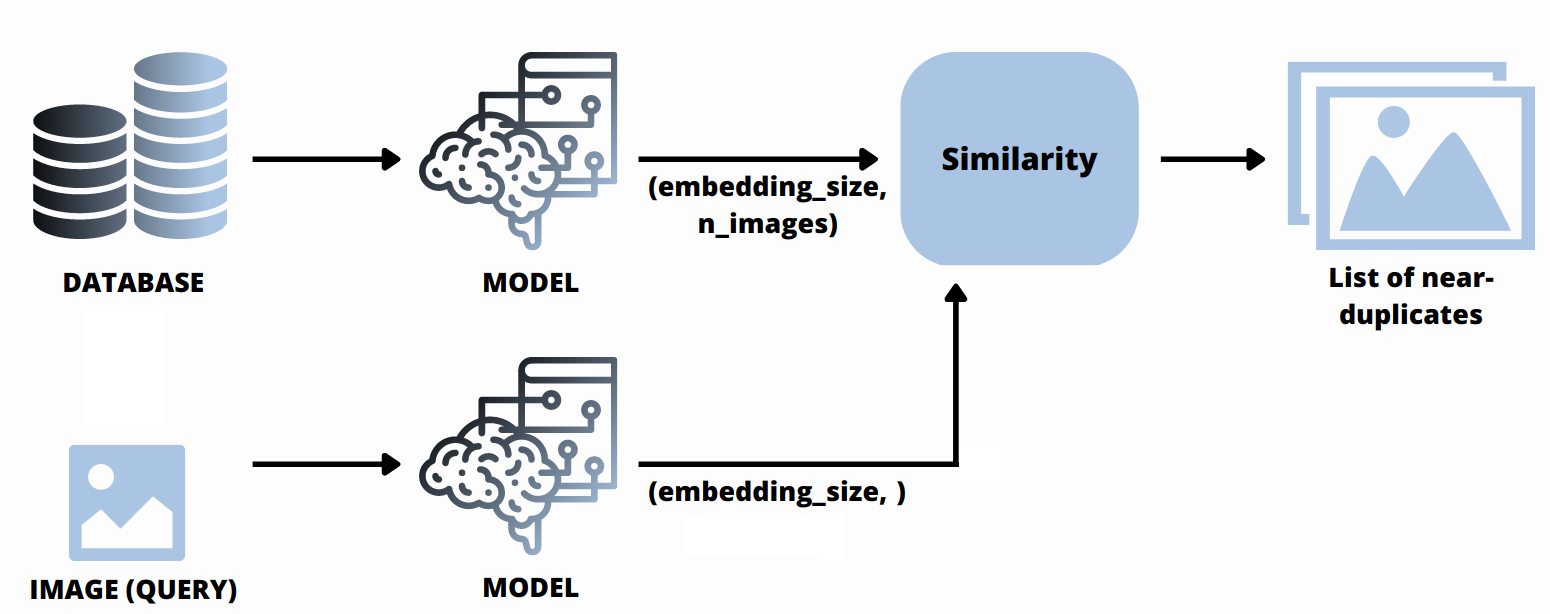}
    \caption{An overview of the near-duplicates detection system.}
    \label{fig:workflow}
\end{figure}

The overview of our near-duplicate detection system is shown in Figure \ref{fig:workflow}. The idea is to predict for each image of the dataset an embedding and then compare through them in order to extract the near-duplicates on the images. In this section, several models to generate embeddings will be discussed. 

\subsection{Hashing Methods}
Traditional hand-crafted methods for detecting near-duplicate images are based on hashing algorithms that extract compact representations of the image and compare them to determine their similarity. Hashing algorithms create fixed-length representations, or hashes, of digital objects like images. For tasks like data deduplication, where the goal is to find and get rid of duplicate copies of data to save space and reduce redundancy, hash algorithms are frequently used. 

The simple Average Hash algorithm reduces the size of the image to $8 \times 8$ square, then reduces its color information by converting it to grayscale, and finally constructs a 64-bit hash where each bit is simply set based on whether the color value of a given pixel is above or below the mean. Average Hash is very fast to compute and robust against scale and aspect ratio changes, brightness/contrast modifications, or color changes. More robust algorithms extend this simple idea using other block-based statistics or the discrete cosine transform (DCT) to further reduce the frequencies of the image \cite{ref_hash}. In this paper, we use two different hashing algorithms' implementations\footnote{\url{https://docs.opencv.org/3.4/d4/d93/group__img__hash.html}} as baselines: Perceptual Hashing and Block Mean Hashing. 

The ability of hashing algorithms to be quick and efficient is one of their benefits. Hash algorithms are suitable for tasks like data deduplication where speed is important because they are built to generate a hash quickly and with minimal computational resources. Another benefit is that, in comparison to deep learning, hash algorithms can be fairly straightforward and simple to implement. They may therefore be a good choice for applications where simplicity and ease of use are crucial factors.

However, there are some drawbacks as well. They might not always be as accurate as alternative learning-based methods, particularly for images that are very similar but not identical. Hashing algorithms may struggle to distinguish minute differences between similar images because they are built to find exact duplicates. Hashing algorithms' sensitivity to changes in the input data is another drawback. For instance, even if an image is only slightly altered or its size is reduced, the resulting hash may differ greatly from the original hash.

\subsection{Deep Learning-based Methods}

The current state of the art approach to compute image similarity involves using deep neural networks. Typical approaches include using off-the-shelf representations from pre-trained models \cite{off_the_shelf}, transfer learning \cite{transfer_learning}, and image retrieval models~\cite{image_retrieval}. In this paper we consider the use of two popular computer vision architectures: ResNet convolutional neural networks (CNN)~\cite{resnet}, and Vision Transformers (ViTs)~\cite{vit}. For both cases we consider supervised and self-supervised training strategies.

\subsubsection{Supervised Learning}

In order to train a model to predict a certain output given a collection of inputs, supervised learning approaches need labeled data. Training a model to identify whether a picture is a near duplicate or not is one way to find near-duplicate photos using supervised learning. The features of the photos may then be retrieved and fed into a deep learning model, such as a convolutional neural network (CNN), using this method. After training, the model may be used to categorize the photos based on their attributes.

To train our near-duplicate image detection models with supervised learning we consider the following two loss functions:

\begin{itemize}
    \item Categorical Cross-Entropy loss: in this case we consider each sub-set of annotated near-duplicate images as a class, and compute $-\sum_{c=1}^My_{i,c}\log(p_{i,c})$, where $M$ is the number of classes, $y$ is a binary indicator (0 or 1) for the correct class label being $c$, and $p$ is the predicted probability for class $c$.
    \item Triplet loss: in this case the model is trained with triplets of images $\{x^a, x^p, x^n\}$ to make the distance low between the embeddings of anchor image $f(x^a)$ and positive image $f(x^p)$ (near-duplicate), while the distance between the embeddings of anchor image   $f(x^a)$ and negative image  $f(x^n)$ (non near-duplicate) is kept large: $\sum _{i}^N [ \left \| f(x^a_i) - f(x^p_i)  \right \|_2^2 - \left \| f(x^a_i) - f(x^n_i) \right \|_2^2 + \alpha]$, where $\alpha$ is the margin parameter.

\end{itemize}

\subsubsection{Self-Supervised Learning}

Thanks to self-supervised learning, models can be trained into large sets of unlabelled data, using the data itself as supervision. The objective is to learn useful visual features from the data without any manual annotations. This is especially useful in scenarios such as the real use case that we study in this article, since normally the new collections that arrive at a document archive manager to be cataloged lack annotations. What distinguishes our use case from self-supervised methods per se is that here we have the data set on which we want to apply our duplicate detection model, and therefore self-supervised learning techniques are a tool of ideal transductive learning.

Two popular self-supervised models will be compared in this paper for our use case scenario: SimCLR (Simple Contrastive Learning)~\cite{ref_lncs1_simclr} and Masked Autoencoders (MAE)~\cite{ref_lncs1_mae}.

In SimCLR, a neural network is trained to determine whether a pair of pictures is a positive pair (i.e., they are different views of the same image) or a negative pair (i.e., they are different images). A contrastive loss function is used to train the model, which motivates it to create identical representations for positive pairings and dissimilar representations for negative pairs. 
\begin{figure}[h]
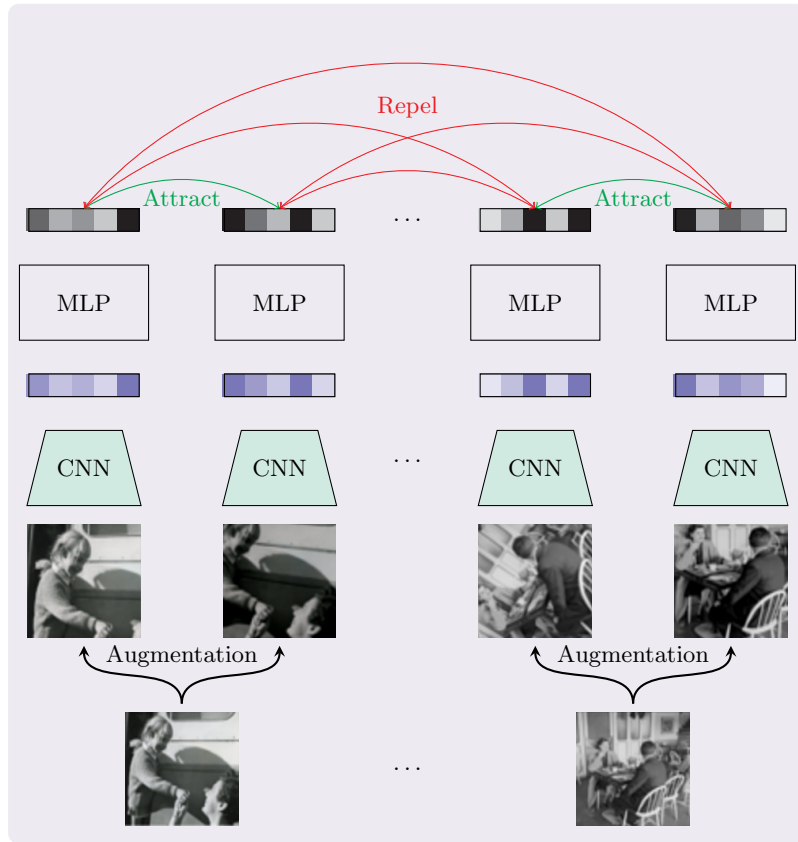

    \centering
    \include{figure_simclr}
    \caption{An illustration of the SimCLR training process.}
    \label{fig:simclr}
\end{figure}

As illustrated in Figure \ref{fig:simclr}, given an image $x$ the first step is to obtain two views of the image ($x_1$ and $x_2$) by applying data augmentation techniques. Then, both views are processed by the visual feature extraction model (e.g. a CNN + MLP) to obtain their embedding representations ($z_1$ and $z_2$). The loss function for a positive pair of examples ($i$, $j$) is defined as:

\begin{equation}
    l_{i,j} = - \log \frac{\exp sim(z_i, z_j) / \tau }{\sum_{k=1}^{2N} \mathbb{1}_{k \neq 1} \exp sim(z_i, z_k) / \tau }
\end{equation}

\noindent where $N$ is the number of images in a given mini-batch, $sim(z_i, z_j)$ is the dot product between $\ell_2$ normalized $z_i$ and $z_j$, $\tau$ is a temperature parameter, and $\mathbb{1}_{k \neq 1} \in \{0, 1\}$ is an indicator function evaluating to 1 iff $k \neq i$.

\begin{figure}[t]
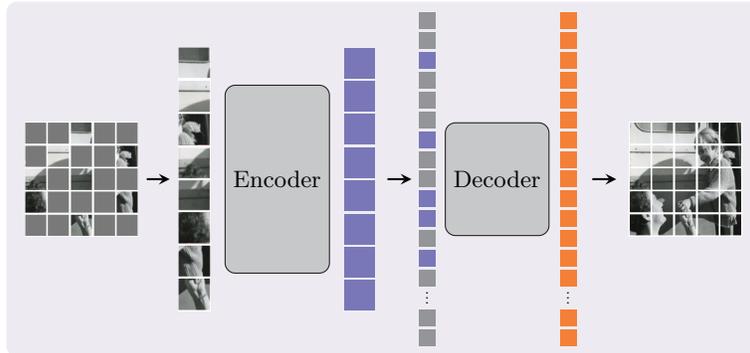

    \centering
    \include{figure_mae}
    \caption{An illustration of the Masked Autoencoders' workflow.}
    \label{fig:masked_autoencoders}
\end{figure}

Masked Autoencoders is another state of the art self-supervised learning technique in which an autoencoder is trained to rebuild an input picture using a portion of its original pixels. A division into non-overlapping patches is created, and then a subset of these patches is randomly sampled and masked so that the model has no access to them. The model is trained to reconstruct the masked image from the unmasked pixels as shown in Figure \ref{fig:masked_autoencoders}, and the difference between the input picture and the reconstructed image is used as a supervisory signal.  

The encoder is a ViT that is applied only to visible (not masked) patches and the decoder has to predict all the tokens (visible and masked). The model is trained with a loss function that computes the mean squared error (MSE) between the reconstructed and original image in the pixel space.

\section{Experiments}

\subsection{Datasets}
\begin{figure}
    \centering
    \includegraphics[width=\textwidth]{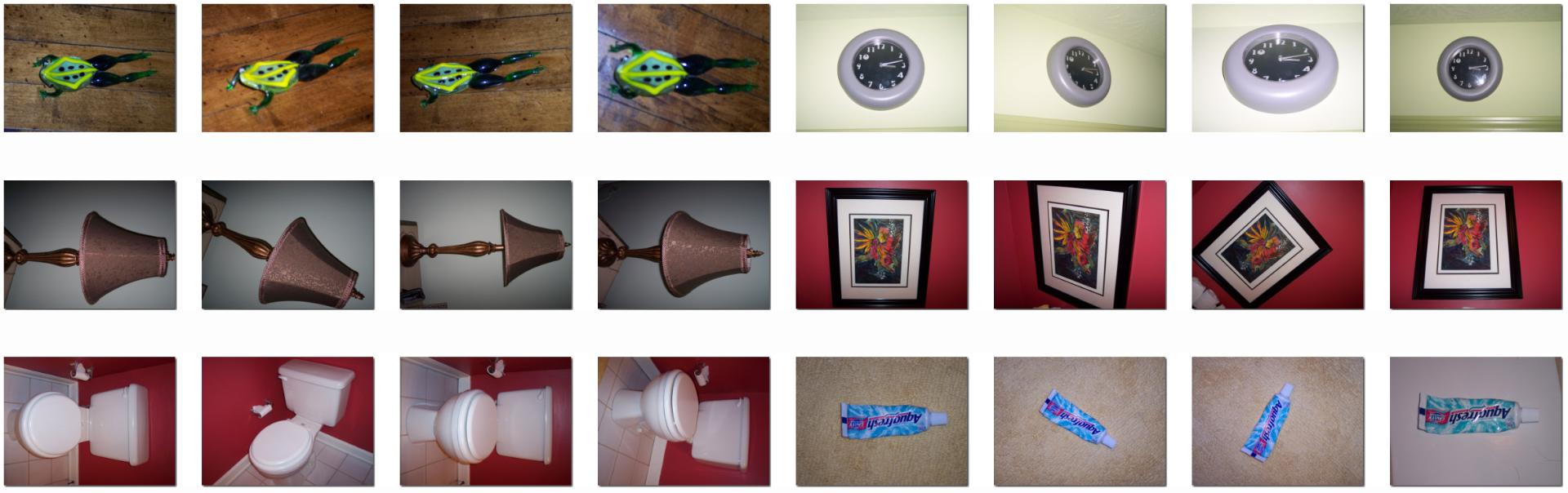}
    \caption{Examples of near-duplicate images from the UKBench dataset. Each image query has 3 near-duplicates that correspond to different points of view, camera rotations, illumination changes, etc.}
    \label{fig:ukbench}
\end{figure}

Two datasets with similar dimensions have been used in our experiments, one is a well-known public benchmark dataset for near-duplicate image detection and the other is a confidential dataset that represents our real use case scenario:

\noindent $\cdot$ \textbf{UKBench} \cite{ref_url_ukbench}: contains 10.200 images, including indoor and outdoor scenes, objects, and people, and is designed to test the ability of retrieval systems to find visually similar images. For each image, 3 near-duplicates are provided (i.e. different points of view, camera rotations, etc.) as shown in Figure~\ref{fig:ukbench}. The dataset is divided into 60/20/20 for train, validation, and test sets. We would like to note that, as seen in Figures 5 and 7, some of the images in this data set contain textual information, either in the form of scene text or handwritten text.
    
\noindent $\cdot$ \textbf{In-house private dataset}: consists of 41.039 images of portraits and artworks. Several views of the same people or similar artworks are provided. In addition, 1014 manual annotations of near-duplicates are given. These annotations contain a query image and one near-duplicate for this specific query. It is important to notice here that the annotations for this dataset are incomplete, we only have one near-duplicate annotation for each of those 1014 queries but other (many) near-duplicate images might exist for each of them. Images from this data set cannot be displayed here due to copyright issues. To get an idea of the type of images in this collection we refer the reader to Figure \ref{fig:teaser} (a) where some examples of public images that are similar to the private data set are shown.

\subsection{Implementation details}

In all our experiments we make use of two different base architectures: a ResNet50 CNN and a ViT-L/16 Transformer. In both cases we use models pre-trained on the ImageNet \cite{imagenet} dataset and fine-tuned on the target datasets. In both the supervised and self-supervised learning settings the models are trained under the same hyperparameters: learning rate of $3 \cdot 10^{-4}$ and Adam optimizer, decaying the learning rate of each parameter group by 0.1 every 8 epochs.

For data augmentation in the SimCLR model, we apply a set of 5 transformations following the original SimCLR setup: random horizontal flip, crop-and-resize, color distortion, random grayscale, and Gaussian blur. The masking ratio for the MAE model is set to 75\%.

\subsection{Metrics}
In order to evaluate the performance of our near-duplicate images detection system we make use of the following metrics:

\noindent $\cdot$ \textbf{Precision@K}: Precision@K is defined as the percentage of near-duplicate items retrieved among the top-K retrieved items. 

\noindent $\cdot$ \textbf{Mean Average Precision (mAP)}: mAP summarizes the overall performance of an image retrieval system of near-duplicates. It represents the mean of the precision values for each query.

In our experiments, we use mAP@4 for the UKBench dataset, since we know that all queries have exactly four near-duplicate images in the retrieval set. There are actually three near duplicates of each query image, but the query image itself is also on the retrieval set, which means there are four relevant images per query. On the other hand, in the private dataset, we evaluate our models using Precision@K for different values of K. This is due to the fact that the dataset is only partially annotated and therefore it is much more difficult to interpret the performance of the models from a single mAP metric value.

\subsection{Results}

Table \ref{table:UKBench} shows the results for different models on the UKBench dataset when using inductive training, i.e. when they are trained/fine-tuned only with the training data. 

\begin{table}[h]
\centering
\caption{Overview of the results for the UKBench dataset.}
\label{table:UKBench}
\begin{tabular}{|c|cc|cc|c|}
\hline
\multirow{2}{*}{\textbf{Model}} & \multicolumn{2}{c|}{\textbf{Pre-training}} & \multicolumn{2}{c|}{\textbf{Fine-tuning}} & \textbf{Metrics} \\ \cline{2-6} 
                                & dataset                  & loss            & dataset           & loss                  & mAP@4            \\ \hline
pHash                           & -                        & -               & -                 & -                     & 0.263            \\
BlockMeanHash                   & -                        & -               & -                 & -                     & 0.321            \\ \hline
ResNet50                        & -                        & -               & UKBench           & C. Entropy            & 0.765             \\
ResNet50                        & ImageNet                 & C. Entropy      & UKBench           & C. Entropy            & \textbf{0.943}            \\
ResNet50                        & ImageNet                 & C. Entropy      & UKBench           & Triplet               & 0.931            \\
ViT-L-16                        & ImageNet                 & C. Entropy      & UKBench           & C. Entropy            & 0.898            \\ \hline
MAE (ViT-L-16)                  & -                        & -               & UKBench           & MSE                   & 0.723             \\ 
MAE (ViT-L-16)                  & ImageNet                 & C. Entropy      & UKBench           & MSE                   & 0.810            \\ 
SimCLR(ResNet50)                & ImageNet                 & C. Entropy      & UKBench           & Contrastive           & 0.727            \\
SimCLR(ViT-L-16)                & ImageNet                 & C. Entropy      & UKBench           & Contrastive           & 0.806            \\ \hline
%CLIP (ViT-B-32)                 & Openai                   & Contrastive     & -                 & -                     & 0.927            \\
%CLIP (ViT-L-14)                 & Laion2b\_s32b\_b82k      & Contrastive     & -                 & -                     & \textbf{0.974}   \\ \hline
\end{tabular}
\end{table}

For the supervised learning setting we evaluate a ResNet50 model trained from scratch, two ResNet50 models pre-trained on ImageNet and fine-tuned on the UKBench dataset (with cross-entropy and triplet losses respectively), and a ViT-L/16 also pre-trained on ImageNet and fine-tuned on the UKBench dataset with cross-entropy. For the self-supervised learning setting, we evaluate two MAEs with ViT-L/16 (one pre-trained on ImageNet and the other trained from scratch), and two models trained with SimCLR: SimCLR(ResNet50) and SimCLR(ViT-L-16). We also show the results of hand-crafted hashing algorithms for comparison.

We appreciate that hashing techniques yield poor outcomes and the best results are obtained with supervised learning. The ResNet50 is working better than the Visual Transformer, which can be explained by the fact that the training dataset is quite small. Self-supervised models exhibit promising outcomes that lag behind the best results of supervised models. At this point, it is clear that supervised learning is the best choice when training data is available for the target domain. Moreover, pre-training on ImageNet is always beneficial for all models.

In the next experiment, we compare the performance of the self-supervised MAE model in the inductive and transductive scenarios for the UKBench dataset. Somehow, what we want is to check what would happen if we didn't have training data annotated for UKBench and therefore we couldn't use supervised learning. This is in fact the scenario we are faced with in our private dataset, for which we have no annotated training data. 

As shown in Table \ref{table:UKBench_type_of_learning} when transductive learning is used in self-supervised MAE models, outcomes are improved significantly. This is because transductive learning allows the model to learn representations directly from the retrieval set, rather than relying solely on generalization to new images. By learning from the retrieval set, the model is able to identify similarities between pairs of images more effectively, resulting in improved performance. Note that the inductive learning results are the same as in Table \ref{table:UKBench} for MAE (ViT-L-16). 

\begin{table}[h]
\centering
\caption{Overview of the results when changing the type of learning used for the UKBench dataset}
\label{table:UKBench_type_of_learning}
\begin{tabular}{|c|c|c|}
\hline
\textbf{Model}                  & \textbf{Partition in which the model is trained} & \textbf{mAP@4} \\ \hline
\multirow{3}{*}{MAE (ViT-L-16)} & Train (inductive)                                & 0.810          \\ \cline{2-3} 
                                & Train and test                                   & 0.824          \\ \cline{2-3} 
                                & Test (transductive)                              & \textbf{0.843} \\ \hline
\end{tabular}
\end{table}

Finally, in Table \ref{table:in-house-dataset} we show the performance comparison of different models in our internal private dataset using transductive learning. As discussed before in this dataset there are not enough annotations to train supervised models. The few available annotations are only used for evaluation purposes, and precision is computed for the top-K retrieved items. The pre-training and fine-tuning data in Table \ref{table:in-house-dataset} are identical to those in Table \ref{table:UKBench}, but omitted here to prevent the table from being too large.

Both Table \ref{table:UKBench} and \ref{table:in-house-dataset} exhibit some correlations when comparing self-supervised models. Models trained with MAEs perform better than the ones trained with SimCLR. ViT-L/16 outperforms ResNet50 in this setting. Overall, results in Table \ref{table:in-house-dataset} are lower than in the UKBench dataset because our in-house private dataset is only partially annotated, many near-duplicates for the same image are present but only one is actually labeled. 

\begin{table}[h]
\centering
\caption{Overview of the results for the in-house private dataset.}
\label{table:in-house-dataset}
\begin{tabular}{|c|cccc|}
\hline
\multirow{2}{*}{\textbf{Model}} & \multicolumn{4}{c|}{\textbf{Metrics}}                                                                                            \\ \cline{2-5} 
                                & \multicolumn{1}{c|}{Precision@1}    & \multicolumn{1}{c|}{Precision@5}    & \multicolumn{1}{c|}{Precision@10}   & Precision@50   \\ \hline
pHash                           & \multicolumn{1}{c|}{0.004}          & \multicolumn{1}{c|}{0.007}          & \multicolumn{1}{c|}{0.011}          & 0.024          \\
BlockMeanHash                   & \multicolumn{1}{c|}{0.015}          & \multicolumn{1}{c|}{0.027}          & \multicolumn{1}{c|}{0.045}          & 0.091          \\ \hline
SimCLR(ResNet50)                & \multicolumn{1}{c|}{0.041}          & \multicolumn{1}{c|}{0.113}          & \multicolumn{1}{c|}{0.165}          & 0.279          \\
SimCLR(ViT-L-16)                & \multicolumn{1}{c|}{0.052}          & \multicolumn{1}{c|}{0.142}          & \multicolumn{1}{c|}{0.198}          & 0.341          \\
MAE (ViT-L-16)                  & \multicolumn{1}{c|}{0.062}          & \multicolumn{1}{c|}{\textbf{0.156}} & \multicolumn{1}{c|}{\textbf{0.218}} & 0.337          \\ \hline
%CLIP (ViT-B-32)                 & \multicolumn{1}{c|}{\textbf{0.066}} & \multicolumn{1}{c|}{0.147}          & \multicolumn{1}{c|}{0.215}          & 0.370          \\
%CLIP (ViT-L-14)                 & \multicolumn{1}{c|}{\textbf{0.066}} & \multicolumn{1}{c|}{0.148}          & \multicolumn{1}{c|}{0.212}          & \textbf{0.389} \\ \hline
\end{tabular}
\end{table}

\begin{figure}[t]
    \centering
    %\includegraphics[width=0.118\textwidth]{img/qualitatives/image1/query.png}
    % \\ Query \\
    \vspace{2.5mm}
    \includegraphics[width=0.118\textwidth]{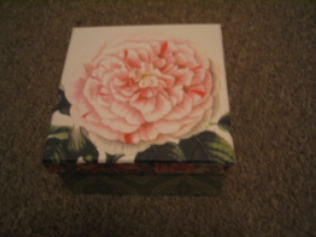}
    \includegraphics[width=0.118\textwidth]{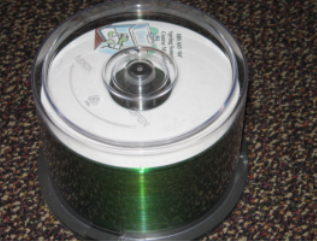}
    \includegraphics[width=0.118\textwidth]{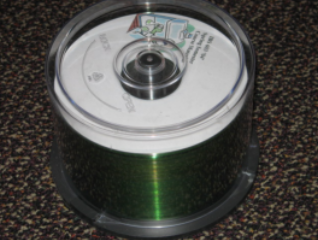}
    \includegraphics[width=0.118\textwidth]{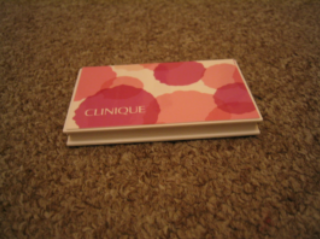}
    \includegraphics[width=0.118\textwidth]{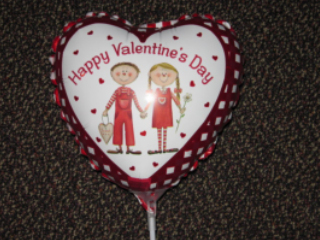}
    \includegraphics[width=0.118\textwidth]{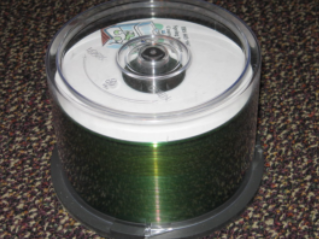}
    \includegraphics[width=0.118\textwidth]{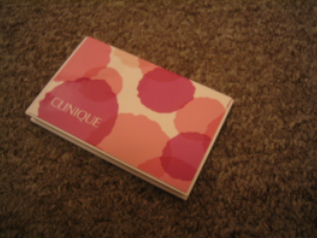}
    \includegraphics[width=0.118\textwidth]{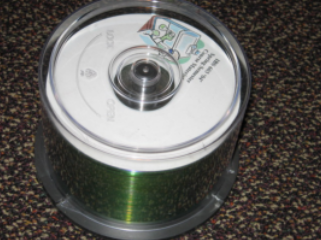}
    \\ (a) \\
    \includegraphics[width=0.118\textwidth]{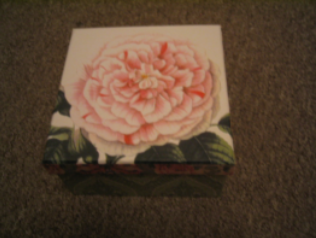}
    \includegraphics[width=0.118\textwidth]{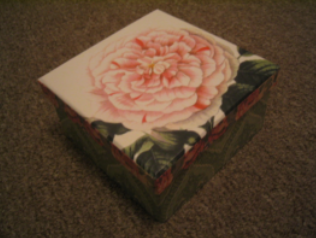}
    \includegraphics[width=0.118\textwidth]{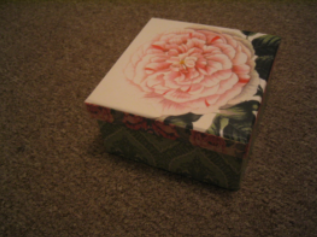}
    \includegraphics[width=0.118\textwidth]{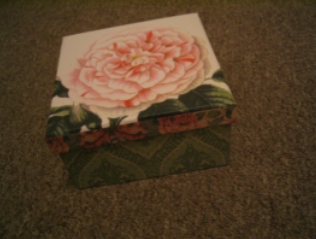}
    \includegraphics[width=0.118\textwidth]{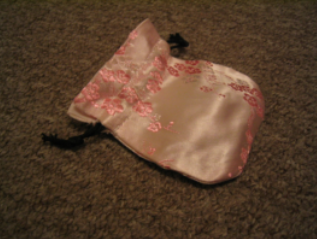}
    \includegraphics[width=0.118\textwidth]{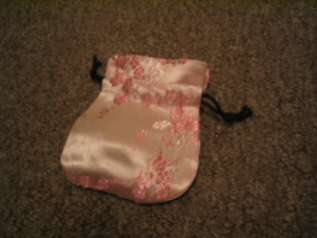}
    \includegraphics[width=0.118\textwidth]{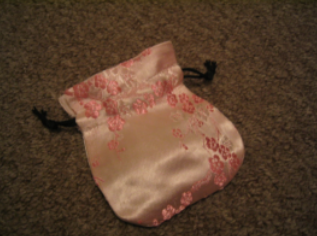}
    \includegraphics[width=0.118\textwidth]{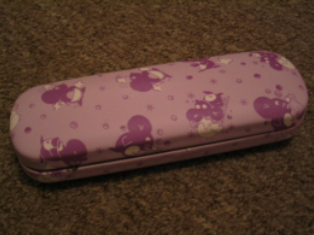}
    \\ (b) \\
    \includegraphics[width=0.118\textwidth]{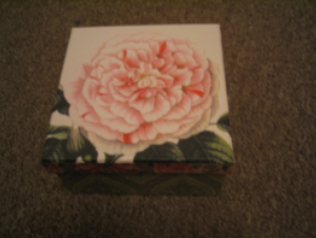}
    \includegraphics[width=0.118\textwidth]{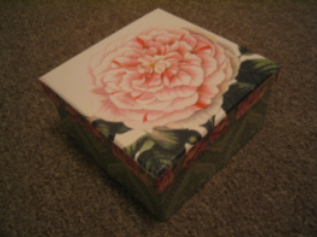}
    \includegraphics[width=0.118\textwidth]{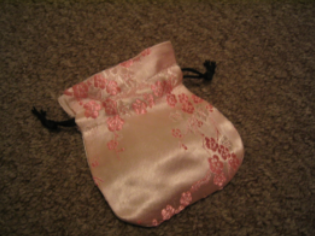}
    \includegraphics[width=0.118\textwidth]{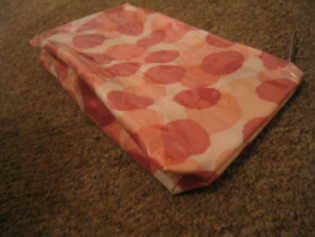}
    \includegraphics[width=0.118\textwidth]{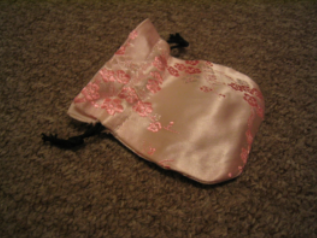}
    \includegraphics[width=0.118\textwidth]{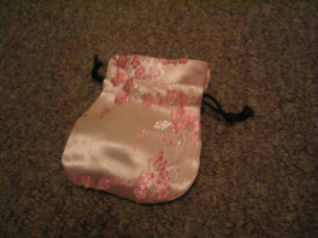}
    \includegraphics[width=0.118\textwidth]{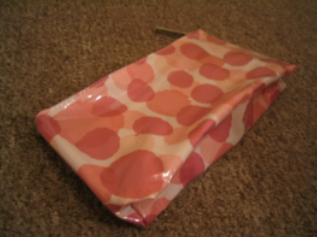}
    \includegraphics[width=0.118\textwidth]{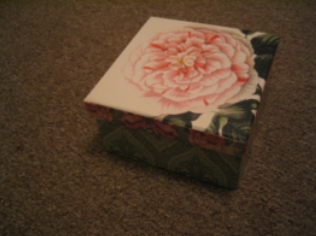}
    \\ (c) \\
%    \includegraphics[width=0.118\textwidth]{img/qualitatives/image1/clip_1.png}
%    \includegraphics[width=0.118\textwidth]{img/qualitatives/image1/clip_2.png}
%    \includegraphics[width=0.118\textwidth]{img/qualitatives/image1/clip_3.png}
%    \includegraphics[width=0.118\textwidth]{img/qualitatives/image1/clip_4.png}
%    \includegraphics[width=0.118\textwidth]{img/qualitatives/image1/clip_5.png}
%    \includegraphics[width=0.118\textwidth]{img/qualitatives/image1/clip_6.png}
%    \includegraphics[width=0.118\textwidth]{img/qualitatives/image1/clip_7.png}
%    \includegraphics[width=0.118\textwidth]{img/qualitatives/image1/clip_8.png}
%    \\ (d) \\
    \caption{Examples of qualitative results on the UKBench dataset for a given query image. We show the top-8 retrieved images ordered by closest distance from left to right for: (a) Block mean hashing, (b) Supervised learning (ResNet50), and (c) Self-supervised learning (MAE). In all cases, the left-most image is both the query and the first retrieved image.}% and (d) zero-shot learning: CLIP}
    \label{fig:qualitatives1}
\end{figure}

\begin{figure}[t]
    \centering
    %\includegraphics[width=0.118\textwidth]{img/qualitatives/image1/query.png}
    % \\ Query \\
    \vspace{2.5mm}
    %\includegraphics[width=0.118\textwidth]{img/qualitatives/image2/query.png}
    %\\ Query \\
    %\vspace{2.5mm}
    \includegraphics[width=0.118\textwidth]{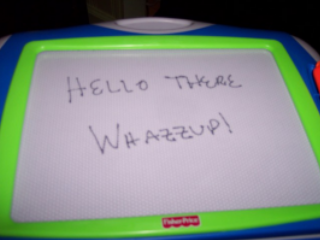}
    \includegraphics[width=0.118\textwidth]{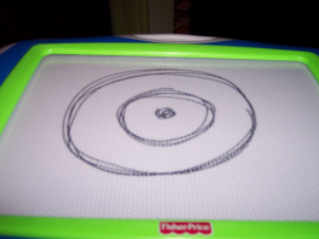}
    \includegraphics[width=0.118\textwidth]{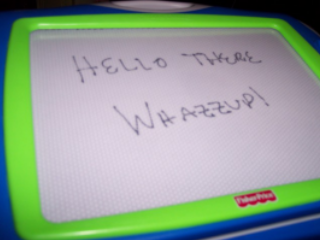}
    \includegraphics[width=0.118\textwidth]{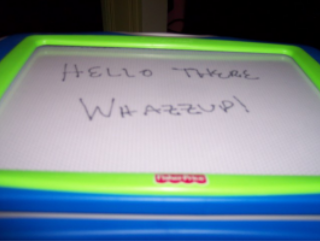}
    \includegraphics[width=0.118\textwidth]{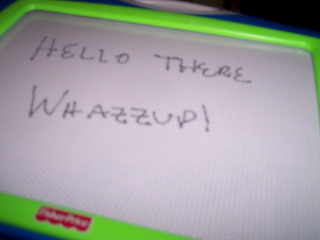}
    \includegraphics[width=0.118\textwidth]{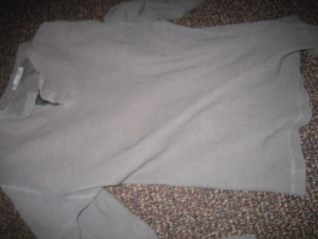}
    \includegraphics[width=0.118\textwidth]{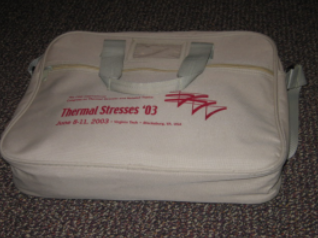}
    \includegraphics[width=0.118\textwidth]{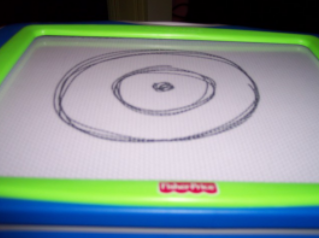}
    \\ (a) \\
    \includegraphics[width=0.118\textwidth]{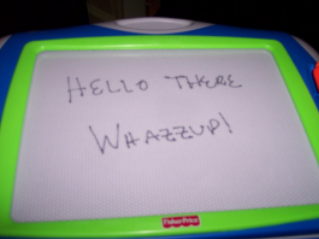}
    \includegraphics[width=0.118\textwidth]{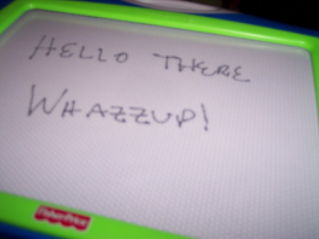}
    \includegraphics[width=0.118\textwidth]{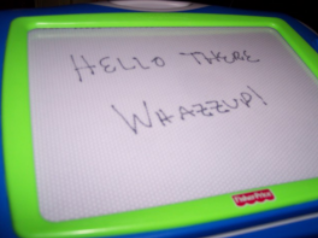}
    \includegraphics[width=0.118\textwidth]{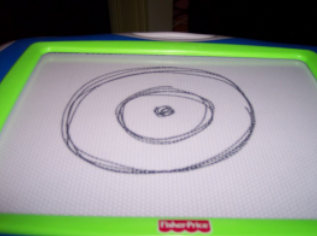}
    \includegraphics[width=0.118\textwidth]{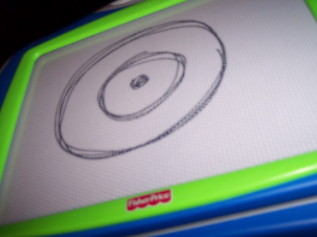}
    \includegraphics[width=0.118\textwidth]{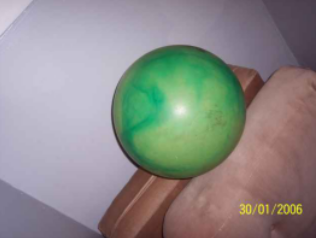}
    \includegraphics[width=0.118\textwidth]{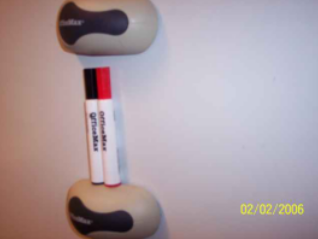}
    \includegraphics[width=0.118\textwidth]{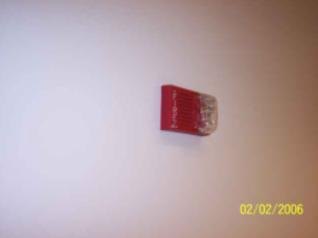}
    \\ (b) \\
    \includegraphics[width=0.118\textwidth]{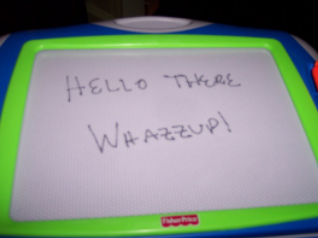}
    \includegraphics[width=0.118\textwidth]{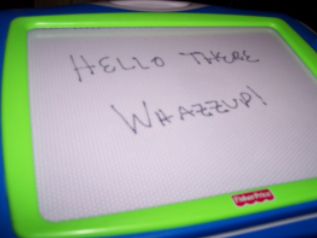}
    \includegraphics[width=0.118\textwidth]{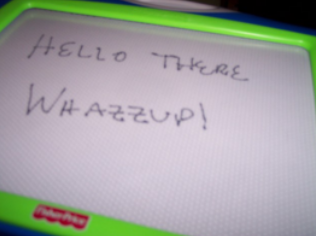}
    \includegraphics[width=0.118\textwidth]{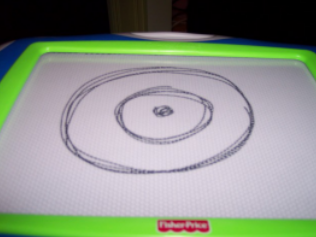}
    \includegraphics[width=0.118\textwidth]{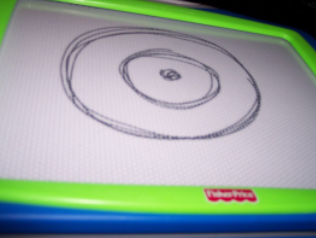}
    \includegraphics[width=0.118\textwidth]{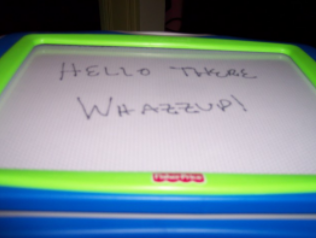}
    \includegraphics[width=0.118\textwidth]{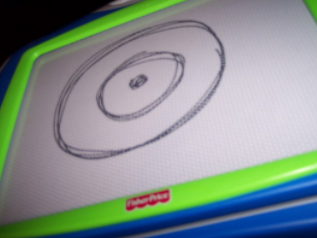}
    \includegraphics[width=0.118\textwidth]{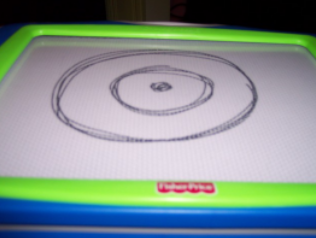}
    \\ (c) \\
%    \includegraphics[width=0.118\textwidth]{img/qualitatives/image2/clip_1.png}
%    \includegraphics[width=0.118\textwidth]{img/qualitatives/image2/clip_2.png}
%    \includegraphics[width=0.118\textwidth]{img/qualitatives/image2/clip_3.png}
%    \includegraphics[width=0.118\textwidth]{img/qualitatives/image2/clip_4.png}
%    \includegraphics[width=0.118\textwidth]{img/qualitatives/image2/clip_5.png}
%    \includegraphics[width=0.118\textwidth]{img/qualitatives/image2/clip_6.png}
%    \includegraphics[width=0.118\textwidth]{img/qualitatives/image2/clip_7.png}
%    \includegraphics[width=0.118\textwidth]{img/qualitatives/image2/clip_8.png}
%    \\ (d) \\
    \caption{Examples of qualitative results on the UKBench dataset for a given query image. We show the top-8 retrieved images ordered by closest distance from left to right for: (a) Block mean hashing, (b) Supervised learning (ResNet50), and (c) Self-supervised learning (MAE). In all cases, the left-most image is both the query and the first retrieved image. Note that in this example the query contains handwritten text and, as can be appreciated, the MAE model has no problem finding its near duplicates.}% and (d) zero-shot learning: CLIP}
    \label{fig:qualitatives2}
\end{figure}

In Figures \ref{fig:qualitatives1} and \ref{fig:qualitatives2}  we present qualitative results of different methods for two given queries of the UKBench dataset. We appreciate that qualitative and quantitative results are consistent: hashing is the method that yields subpar outcomes while deep learning models trained with supervised or self-supervised learning strategies exhibit comparable results. In Figure \ref{fig:qualitatives2} we show an example where the query image contains handwritten text, demonstrating the potential use of the studied techniques in document collections as well as photo collections.


\section{Conclusions}
The present study provides a comprehensive evaluation of various models for embedding images in order to detect near-duplicates. The results demonstrate that training a self-supervised model is a suitable approach when labels are missing, as it enables the learning of the entire dataset's characteristics without the need for manual labeling. The proposed transductive learning approach, which involves fine-tuning a deep neural network on the target retrieval set with self-supervision, shows promising results in terms of retrieval metrics, outperforming the inductive learning baselines.

While synthetic data can be a useful tool for pre-training models, the focus in this paper has been on fine-tuning pre-trained models, using transductive learning on real data from the retrieval collection. While the use of GANs or Stable Diffusion can increase the size and diversity of the pre-training set, it has been out of the scope of this paper and could be an interesting approach for additional investigation.

Future research could also investigate the impact of using ViT foundational models pre-trained on larger datasets, fine-tuned using self-supervised learning, on the near-duplicate detection problem in the specific context of archival work. An other interesting

\section*{Acknowledgements}
This work has been supported by the ACCIO INNOTEC 2021 project Coeli-IA (ACE034/21/000084), and the CERCA Programme / Generalitat de Catalunya. Lluis Gomez is funded by the Ramon y Cajal research fellowship RYC2020-030777-I / AEI / 10.13039/501100011033.

\bibliographystyle{splncs04}

\begin{thebibliography}{8}
\bibitem{ref_dl_book}
{Goodfellow, Ian, Yoshua Bengio, and Aaron Courville. Deep learning. MIT press, 2016.}

\bibitem{ref_transductive}
{Joachims, Thorsten. "Transductive learning via spectral graph partitioning." Proceedings of the 20th international conference on machine learning (ICML-03). 2003.}

\bibitem{ref_lncs1_survey_nd}
Thyagharajan, K. K., and G. Kalaiarasi. "A review on near-duplicate detection of images using computer vision techniques." Archives of Computational Methods in Engineering 28.3 (2021): 897-916.

\bibitem{ref_lncs1_nd4}
Erin Liong, Venice, et al. "Deep hashing for compact binary codes learning." Proceedings of the IEEE conference on computer vision and pattern recognition. 2015.

\bibitem{ref_lncs1_nd5}
Liu, Haomiao, et al. "Deep supervised hashing for fast image retrieval." Proceedings of the IEEE conference on computer vision and pattern recognition. 2016.

\bibitem{ref_lncs1_nd6}
Wu, Dayan, et al. "Deep supervised hashing for multi-label and large-scale image retrieval." Proceedings of the 2017 ACM on International Conference on Multimedia Retrieval. 2017.

\bibitem{ref_lncs1_nd7}
Zhao, Fang, et al. "Deep semantic ranking based hashing for multi-label image retrieval." Proceedings of the IEEE conference on computer vision and pattern recognition. 2015.

\bibitem{ref_lncs1_nd1} 
Zhou, Zhili, et al. "Near-duplicate image detection system using coarse-to-fine matching scheme based on global and local CNN features." Mathematics 8.4 (2020): 644.

\bibitem{ref_lncs1_nd2}
Morra, Lia, and Fabrizio Lamberti. "Benchmarking unsupervised near-duplicate image detection." Expert Systems with Applications 135 (2019): 313-326.

\bibitem{ref_lncs1_nd3}
Zhang, Yi, et al. "Single-and cross-modality near duplicate image pairs detection via spatial transformer comparing CNN." Sensors 21.1 (2021): 255.

\bibitem{ref_lncs1_nd8}
{Chum, Ondrej, James Philbin, and Andrew Zisserman. "Near duplicate image detection: Min-hash and TF-IDF weighting." Bmvc. Vol. 810. 2008.}

\bibitem{ref_lncs1_nd9}
{Dong, Wei, et al. "High-confidence near-duplicate image detection." Proceedings of the 2nd acm international conference on multimedia retrieval. 2012.}

\bibitem{ref_lncs1_nd10}
{He, Bing, et al. "Part-regularized near-duplicate vehicle re-identification." Proceedings of the IEEE/CVF Conference on Computer Vision and Pattern Recognition. 2019.}

\bibitem{ref_hash}
Zauner, Christoph. "Implementation and Benchmarking of Perceptual Image Hash Functions." (2010).

\bibitem{off_the_shelf}
Sharif Razavian, Ali, et al. "CNN features off-the-shelf: an astounding baseline for recognition." Proceedings of the IEEE conference on computer vision and pattern recognition workshops. 2014.

\bibitem{transfer_learning}
Yosinski, Jason, et al. "How transferable are features in deep neural networks?." Advances in neural information processing systems 27 (2014).

\bibitem{image_retrieval}
Dubey, Shiv Ram. "A decade survey of content based image retrieval using deep learning." IEEE Transactions on Circuits and Systems for Video Technology 32.5 (2021): 2687-2704.

\bibitem{resnet}
He, Kaiming, et al. "Deep residual learning for image recognition." Proceedings of the IEEE conference on computer vision and pattern recognition. 2016.

\bibitem{vit}
Dosovitskiy, Alexey, et al. "An image is worth 16x16 words: Transformers for image recognition at scale." arXiv preprint arXiv:2010.11929 (2020).

\bibitem{ref_lncs1_simclr}
Chen, Ting, et al. "A simple framework for contrastive learning of visual representations." International conference on machine learning. PMLR, 2020.

\bibitem{ref_lncs1_mae}
He, Kaiming, et al. "Masked autoencoders are scalable vision learners." Proceedings of the IEEE/CVF Conference on Computer Vision and Pattern Recognition. 2022.

\bibitem{ref_lncs1_clip}
Radford, Alec, et al. "Learning transferable visual models from natural language supervision." International Conference on Machine Learning. PMLR, 2021.

\bibitem{ref_url_ukbench}
Nister, David, and Henrik Stewenius. "Scalable recognition with a vocabulary tree." Computer vision and pattern recognition, 2006 IEEE computer society conference on. Vol. 2. Ieee, 2006.

\bibitem{imagenet}
Deng, J. et al., 2009. Imagenet: A large-scale hierarchical image database. In 2009 IEEE conference on computer vision and pattern recognition. pp. 248–255.

%Author, F.: Article title. Journal \textbf{2}(5), 99--110 (2016)

%Author, F., Author, S.: Title of a proceedings paper. In: Editor,
%F., Editor, S. (eds.) CONFERENCE 2016, LNCS, vol. 9999, pp. 1--13.
%Springer, Heidelberg (2016). \doi{10.10007/1234567890}

%Author, F., Author, S., Author, T.: Book title. 2nd edn. Publisher,

%Author, A.-B.: Contribution title. In: 9th International Proceedings
%on Proceedings, pp. 1--2. Publisher, Location (2010)

%LNCS Homepage, \url{http://www.springer.com/lncs}. Last accessed 4

\end{thebibliography}

\end{document}